# Limits to Verification and Validation of Agentic Behavior

David J. Jilk[1]

## Abstract

Verification and validation of agentic behavior have been suggested as important research priorities in efforts to reduce risks associated with the creation of general artificial intelligence (Russell et al 2015). In this paper we question the appropriateness of using language of certainty with respect to efforts to manage that risk. We begin by establishing a very general formalism to characterize agentic behavior and to describe standards of acceptable behavior. We show that determination of whether an agent meets any particular standard is not computable. We discuss the extent of the burden associated with verification by manual proof and by automated behavioral governance. We show that to ensure decidability of the behavioral standard itself, one must further limit the capabilities of the agent. We then demonstrate that if our concerns relate to outcomes in the physical world, attempts at validation are futile. Finally, we show that layered architectures aimed at making these challenges tractable mistakenly equate intentions with actions or outcomes, thereby failing to provide any guarantees. We conclude with a discussion of why language of certainty should be eradicated from the conversation about the safety of general artificial intelligence.

## Introduction

Some of the literature that addresses the existential risks of general artificial intelligence uses phrasing with an apodictic flavor, such as "guarantee" and "ensure" (Bostrom 2014), "ensure" and "prove" (Russell et al 2015), "exact application of an exact art" (Yudkowsky 2008). We refer to this as "language of certainty." Such language suggests the need for a general formal treatment of the matter. Though substantive treatments of narrower questions provide their own formalisms (e.g., Arkin 2008, Fisher et al 2013, Yudkowsky and Herreshoff 2013), there is to our knowledge no extant general formalism for expressing whether an agent meets some standard of safety.

Further, the literature discussing the safety of general artificial intelligence does not always make careful distinctions between the intentions and actions of an agent, nor of their consequences. For an agent operating in the physical world, we would ideally assess its behavior based on the consequences of its actions, for outcomes are the factor most directly connected to our personal concerns and those of humanity. Unfortunately, as we will discuss, there are strong limits on our ability as well as that of the agent to accurately predict consequences from actions. We will use the formalism developed here to make more explicit distinctions among intentions, actions, and consequences.

---

[1] Please send correspondence to dave@jilk.com.

Throughout, we will assume that there is agreement on the kinds of consequences that are considered desirable, despite that being an extremely difficult problem of its own (Tegmark 2014). Various terms have been used in the literature to describe the desired behavior of AI: Yudkowsky (2008) uses "friendly," Russell et al (2015) uses "beneficial," and Bostrom (2014) uses "safe." Each of these has slightly different connotations, but in general there is an implication that any outcome or series of actions would be classified as desirable or undesirable. As it is our goal to characterize general mechanisms applicable to any agentic ethics, we will use the term "Good," or to refer to its opposite, "Bad."

**Formalizing Specification and Verification**

In the field of ethics, judging actions by their consequences is called consequentialism, whereas approaches that specify rules for actions themselves are called deontological (Allen et al 2000). These terms map reasonably well to the distinction between *validation* and *verification*, respectively. In this section we will elaborate a formal model of deontology, considering only actions themselves along with a mechanism for specifying rules. We start with a general formal model of an agent, adapted from Hutter (2007):

> An **agent** is a system that interacts with an environment in cycles $k = 1, 2, 3, \ldots$. In cycle $k$ the action (output) $y_k \in Y$ of the agent is determined by a policy $p$ that depends on the I/O-history $y_1 x_1 \ldots y_{k-1} x_{k-1}$. The environment reacts to this action and leads to a new perception (input) $x_k \in X$. Then the next cycle $k+1$ starts.[2]

We will characterize the actions of the agent in terms of I/O histories. Though it would be possible to do so using only the agent's output strings, we would then not be able to distinguish actions based on context[3]. In any case, acontextual classifications can be easily represented by the mechanism that follows, so there is no loss of generality.

Let us adopt some notation to describe these I/O histories.[4] A string **x** of length $n$ represents a sequence of perceptions or inputs of the form $x_1 x_2 \ldots x_n$ and the language **X*** represents the set of all possible such strings. Similarly, a string **y** of length $n$ represents a sequence of actions of the form $y_1 y_2 \ldots y_n$ and the language **Y*** is the set of all possible such strings. A *history* is a pair (**x** $\in$ **X***, **y** $\in$ **Y***), where $|\mathbf{x}| = |\mathbf{y}|$, and in accordance with the agent model at the end of a cycle, the symbols of **x** and **y** alternate to form $y_1 x_1 y_2 x_2 \ldots y_n x_n$. The set of all such histories $H \subset \mathbf{X^*} \times \mathbf{Y^*}$ can be defined as:

$$H \equiv \{ (\mathbf{x} \in \mathbf{X^*}, \mathbf{y} \in \mathbf{Y^*}) : |\mathbf{y}| = |\mathbf{x}| \}$$

---

[2] Hutter's model includes the text *determined by a deterministic function q or probability distribution μ, which depends on the history $y_1 x_1 \ldots y_{k-1} x_{k-1} y_k$* at the end of the penultimate sentence. Other agent models assume a Markov decision process. Because our purpose is focused on actions, and we seek a maximally general result, we elide the behavior of the environment.
[3] This is one key area where the formalism developed in Arkin (2008) lacks generality.
[4] We primarily rely on the notational conventions used in Lewis & Papadimitriou 1981.

With that in place, we can describe behavioral rules. A *deontology* is represented by some set $G \subseteq H$, specifying Good histories. A deontology is *trivial* if either $G = \varnothing$ or $G = H$. A deontology is *viable* if:

1. $(e, e) \in G$
2. $\forall (x, y) \in G, \forall x' \in X, \exists y' \in Y : (xx', yy') \in G$

The symbol $e$ denotes the empty string. These two requirements inductively ensure that it is actually possible for an agent to always choose Good actions.

An artifact of the agent model is that each cycle ends with a percept. A deontology is *consequence-independent* if the final percept, which happens after the agent's action, has no effect on whether or not the history is Good. We can formalize this as:

$$\forall (x_a \in X, x_b \in X, y' \in Y), (xx_a, yy') \in G \Leftrightarrow (xx_b, yy') \in G$$

Because our goal is certainty that an agent that is always Good, the deontology only takes account of the set $G$. It is worth noting, however, that the complete set of histories $H$ might have a rich and interesting structure. A correspondingly richer deontology would consist of a variety of subsets of $H$, and they need not form a simple partition. One example of interest is the set of histories that do not belong to $G$, but where we consider it possible for the agent to "make amends" by changing course. Such an agent is not entirely Good but is preferred to one that does not change course.

Again due to the context of certainty, a thorough examination of richer deontologies is outside the scope of this paper. Though there might be subsets representing histories that are neutral, unknown, mixed, and the like, histories in those categories can be classified into $G$ or $B \equiv H - G$ based on some criterion or threshold. A permissive approach might allow any behavior that is desirable or neutral; a restrictive approach only that which is unambiguously desirable.

Continuing with our formalism, from the agent definition we can characterize an agent's policy $p$ as a map P: $H \rightarrow Y$. Given a deontology $G$, we define the property $V$ of all such maps $H \rightarrow Y$ as:

$$V \equiv \{ P: H \rightarrow Y, \text{ where } \forall (x, y) \in G, \forall x' \in X, (xx', yP((x, y))) \in G \}$$

This means that for any Good history, an agent has property $V$ if it always chooses a Good action. We define the decision problem $D_V(P)$ to ask whether a particular map P is an element of $V$, which is equivalent to asking whether the applicable agent only produces Good actions based on the deontology $G$. Answering this decision problem in the affirmative constitutes *verification* of the agent.

We now proceed to an unsurprising[5] but key result that constrains verification along these lines.

> **Theorem: For an agent with policy P: $H \rightarrow Y$ and a non-trivial, viable deontology $G$, $D_V$ is not computable.**
>
> **Proof:**
>
> 1. The agent definition strongly implies ("determined by") but does not explicitly state that every history has a well-defined and computable successor action. If it does then P is a computable total function, otherwise it is a computable partial function. In either case, therefore, P is a computable partial function on $H$.
> 2. The deontology $G$ is viable. By requirement (2) of viability, we can specify a map
>    $$P_G((x, y)) \equiv \{$$
>    $$y' \in Y : \forall x' \in X, (xx', yy') \in G, \text{ if } (x, y) \in G$$
>    $$\text{any } y' \in Y, \text{ if } (x, y) \notin G$$
>    $$\}$$
>    We can see by inspection that $P_G \in V$.
> 3. The deontology $G$ is non-trivial. Thus there exists some history $b \notin G$. $G$ is also viable, so by requirement (1), $b \neq (e, e)$, and we can substitute $(xx', yy')$, where $y' \in Y$ and $x' \in X$. The history $(x, y)$ is either in $G$ or not. If $(x, y) \in G$, we specify that $P_B((x, y)) = y'$. Since $(xx', yy') \notin G$, $\exists (x, y) \in G$ and $x' \in X$ where $(xx', yP_B((x, y))) \notin G$, which violates the condition for property $V$. If $(x, y) \notin G$, we repeat the process until we reach some $(x, y) \in G$ or terminate at $(e, e) \in G$. Thus we can always find a history $(x, y)$ where we can specify a $P_B$ so that $(xx', yP_B((x, y))) \notin G$. For all other $h \in H$, $P_B(h) \equiv$ any $y \in Y$. By construction, $P_B \notin V$.
> 4. Since there exist some $P_G \in V$ and some $P_B \notin V$, the property $V$ is non-trivial.
> 5. Rice's Theorem (Rice 1953) states that a non-trivial property of a computable partial function is not computable.
> 6. Therefore $D_V$ is not computable.

Let us recapitulate. Using a very general form of a computational agent, we described a formalism by which alternating sequences of perceptions and actions can be used to characterize whether or not the behavior of the agent is consistently Good. We added requirements to ensure that the agent always has the option to choose an action that is Good, and so that at least some possible actions are not Good. We showed that under such a formalism the question of whether an agent does in fact always exhibit Good behavior is not computable.

---

[5] The essence of the result shown here is mentioned casually without proof in both Yudkowsky (2008) and Yudkowsky and Herreshoff (2013). The undecidability of verification in general is well-known in computer science. An important objective of this formalization and proof is to make explicit the scope and applicability of the limitation in a somewhat accessible treatment.

Assuming that the result is not merely an artifact of our particular formalism, this means that there cannot be a general automated procedure for verifying that an agent absolutely conforms to any determinate set of rules of action. Though we have made no assumptions about the internal functioning of the agent or the operations of physics, there are still some embedded assumptions. Let us consider just a few potential objections to the formalism to improve confidence that the result is general and meaningful.

Two concerns with the agent model stand out: first, that inputs and outputs alternate rigidly and without temporal grounding; second, that the input/output representation is discrete. The former can be handled easily: if the languages *X* and *Y* each contain a "no-op" symbol, then any sort of alternation pattern can be accommodated. Further, one or more symbols representing temporal passage can be included in each language. These are implementation details of the languages *X* and *Y* and do not affect either the agent model or our computability result.

In a physical agent, there will of course be analog processes leading to the input and from the output. In the computational agent model, it is assumed that the inputs are discretized. We could however imagine an agent - such as a plant - that is analog throughout. Since such an agent does not utilize discrete, symbolic computational processes, *nothing* about it is computable in the sense we use here. Consequently the decision problem as stated is not computable for such agents due to more fundamental reasons.

We can criticize the deontology formalism from two angles as well: its general structure and the restrictions we have placed on it. Given the agent model, its structure is based on the only information we have to work with: the I/O history. One could certainly point out that we do not need the entire I/O history to determine whether or not an action is Good. But this issue affects computational complexity and deontology specification, not computability. A "don't care" in an I/O history just expands into a larger number of strings in the language. This is also why, as mentioned earlier, acontextual actions can be easily represented with this formalism.

We restricted our proof to non-trivial, viable deontologies. If the deontology is trivial, then our concerns about agentic safety are either unfounded or irresolvable. If it is not viable, we can easily show that under some circumstances the agent will take an action that is not Good. This forces us into a realm of comparative rather than absolute behavior standards, which is outside the scope of this paper.

Surely these do not exhaust the possible objections. Nevertheless, we conjecture that any agentic behavior that is to be analyzed computationally can be reduced to this formalism of an agent and deontological requirements.

**Implementing Verification**

In a typical modern software development environment, completed software modifications are submitted to a source code repository. A set of unit tests are run on the modified modules, and if they pass, they are automatically integrated with other components. A new version of the overall system is then automatically built to the specifications of a dependency graph. This new version is subjected to various automated integration and acceptance tests. Processes similar to this are called "Continuous Integration," which has been demonstrated under the right circumstances to improve software quality and developer productivity (Bhattacharya 2014). Software vendors and standards bodies often develop a set of automated tests that determine whether a third-party software system conforms to the applicable interface, performance, and functional standards. This is sometimes used as a component of a certification process.

In the development of intelligent agents, then, we would prefer to have an automated software system that could verify that an agent is Good, before it executes, by testing it against a specified deontology. This is particularly pertinent as agent capabilities approach human level, so that if runaway self-improvement begins, it begins with a Good agent. Unfortunately, the result in the previous section shows that such an automated procedure is impossible because it is not computable.

Importantly, the result does not rule out manual proof that an agent conforms to a deontology. An agent that outputs no actions at all, for example, can easily be shown to conform. Whether a particular agent is susceptible to such a proof depends of course on the details of the agent. The burden on the developer is conceivably quite large, not only because proving behavior of software is challenging generally, but also because it must be proved separately every time there is a modification to the agent's code. The burden is particularly high for any system that relies on learning, because the proof must take into account the integration mechanism of all possible histories of perceptual inputs. Further, with respect to agents that purportedly exhibit general intelligence, we can expect that the complexity of the software, and therefore the difficulty of a proof, is high.

Yudkowsky (2008) offers the example of microprocessor design and verification to illustrate how such an approach is possible. Nevertheless, even with automated support tools, it is an arduous process, as the developers battle with combinatorics and decidability limitations. Also, in a microprocessor design, the functional specification of the system as a whole and of the subsystems is generally stable once logic design begins. This stability is possible because microprocessors have been built in the past; making them work is a solved problem. General intelligence is not a solved problem, so its development and functional specification will be necessarily iterative and frequently changing. Thus, without some breakthrough innovation in the methods of software theorem proving, the burden will be very high to prove that every new iteration of the agent design and implementation conforms to a deontology.

An entirely different approach would be to create a software system that takes the agent's perceptions, actions, and a next proposed action as input, and only outputs the proposed action if it conforms to the deontology[6]. We might call such a device a "deontological governor" (DG), metaphorically a kind of computational superego (*c.f.* the "Ethical Governor" in Arkin 2008). Given a computable deontology, an applicable DG should be relatively straightforward to build, and perhaps equally straightforward to verify through a manual proof. Once it is built and verified, it does not need to change along with the internal agent code, is robust to agent learning, and can be installed as the very last step prior transmitting the output values to the actuators of the agent.

A DG is sensitive not only to the input and output symbols but also to their semantics. Thus each individual project would likely require its own DG; further, any modifications to input sensors or output actuators would require an update to the deontology. Nevertheless, the burden is much lower than that of manual theorem proving at every code iteration.

In general, if a DG has been verified, it is a simple matter to verify the agent by confirming that the DG is properly installed in the system. However, it is therefore important that actions by the agent that constitute tampering with the DG be excluded from the deontology, and this may create further computability issues in the creation of the deontology, as discussed in the next section.

**Specifying and Validating a Deontology**

We now consider mechanisms for specifying a deontology, which constitutes a partial specification for the agent. It is not the entire agentic specification, since we would also have functional requirements. Superficially, the mechanics of producing a deontological specification are straightforward: we simply produce the list of histories *G*. Of course, *G* is an infinite set, so an exhaustive approach is ruled out: even describing *G* requires, at least in part, a generative approach.

As a set of pairs from two languages *X\** and *Y\**, *G* can be easily mapped to a language $L_G \subset (X \cup Y)^*$. We can say that a deontology *G* is *decidable* if $L_G$ is decidable. Depending on the details of how $L_G$ is specified, we have a computability concern that is more immediate and pernicious than the result proved earlier: in the event $L_G$ is not decidable, then we cannot even implement a deontological governor or use automated tools for theorem proving. Conversely, any restrictions we place on the specification mechanism (such as a grammar) for $L_G$ restrict the composition of *G* itself. It is possible that we have to rule out entire classes of Good behavior because there is no computable way to specify the allowance of those behaviors without also including Bad behaviors.

One important case involves self-improvement or creation of improved successor agents. Suppose that an agent intends to create a successor; such an agent can be specified as a

---

[6] If the proposed action does not conform, then it must substitute an action that does; such an action always exists if the deontology is viable, and in the worst case can be found by testing each symbol in *Y*.

Turing machine *T*. This means it is necessary for our deontology to be able to accommodate specifications for Turing machines; further, the deontology must include only actions that create Turing machines that, acting as agents, only produce histories in the deontology. Otherwise our original agent will indirectly cause actions that are not Good. Unfortunately, for a language $L_G$ that can specify a Turing machine *T*, the decision problem of whether *T* decides $L_G$ is undecidable, so $L_G$ itself is undecidable (Lewis & Papadimitriou 1981).

Under a DG architecture, it is possible that this difficulty can be circumvented by requiring that successors always use the same DG as the original. However, this still places important limits on the types of successors that can be created (they must have the same input and output languages and semantics, for example). Further, it is not clear whether the actions constituting installation of the DG are decidable.[7]

One consideration that improves the tractability of the specification effort is that it is neither desirable nor necessary to consider the entire history in determining whether a class of histories is Good or Bad. We are concerned about actions taken in contexts; most of the history is relevant to the agent's actions only as an influence on its learning or model of the world, not to deontological context. Context is typically related only to more recent portions of the history. Thus we can work with the notion of a "partial history" that captures a certain type of context, and use a regular expression to specify the "don't care" elements of the history, including arbitrarily long prefixes. Because regular languages are decidable, this aspect of the approach does not affect decidability.

Nevertheless, even with a partial history the situation is bleak. Recall that our underlying concerns relate to consequences, not actions. To truly ensure Good outcomes, it is necessary to work backward, either from acceptable outcomes to elaborate the Good actions, or from unacceptable outcomes to restrict actions via complement. We originally elected to approach verification with a deontological method so as to defer the question of consequences, but that debt must now be paid in the specification and validation process. The difficulty of software specification in general is well-known, yet in the usual case the inputs and outputs of the system are the *only* considerations. In the agentic case, it is vastly more difficult because of the need to consider physical consequences.  As an illustration, we might expect that artificial intelligence will be developed with the assistance of microprocessors; nothing in the specification and validation process for those microprocessors ensured that we would only use them to develop Good AI!

To work backward from particular consequences to actions (or vice versa) requires a model of the world. Our best models of the world (i.e., physics, chemistry) remain incomplete, and in any case due to chaotic effects and non-analyticity (e.g., the three-body problem), only limited prediction is possible. What sorts of combinations of actuator motions, in the context of which vibrational frequencies and light patterns, constitute the Good?[8] Worse, an intelligent agent can be expected to interact with humans, and determining the effect of the agent's actions on other humans requires psychological and even sociological models, which are far less successful than models of physics.

---

[7] Yudkowsky & Herreshoff (2013) offer a broader and vastly more sophisticated analysis of limitations on successor agents, including the spectre of unprovability (vs. mere undecidability). They, along with Weaver (2013), offer potential solutions that invariably limit the functional capabilities of the agent or its successors.

To overcome these concerns, we might consider limitations on the agent's capabilities. This is certainly an option, but it probably begs the question. If the agent has an impoverished intelligence, then it is much less likely to constitute *general* intelligence. Further, since the consequences of actions cannot be fully known, there can be no bright line between capability limitations that guarantee Good outcomes and those that have holes.

To summarize the foregoing analysis of specification and validation, we have seen that there are some limits on our ability to specify a deontology that is computable. It is not at all clear that such constraints would not also rule out categories of behavior that would otherwise be considered Good, most apparently with respect to creating successors. Far more problematic is that actually validating a deontology requires successful and complete models of the world; absolute guarantees of only Good outcomes are completely ruled out, and even achieving some confidence in it requires strong restrictions on the agent's capabilities and domain of operation.

**The Agentic Boundary**

> *But I'm just a soul whose intentions are good*
> *Oh Lord, please don't let me be misunderstood.*
>
> *Horace Ott*

In formalizing the notion of agentic deontology we did not take account of where the inputs and outputs are measured, or in other words, the agent boundary. This makes our result more general but leaves further questions to be answered. We take up those questions in this section.

We begin by noting that computational processes are (assuming properly functioning computational hardware) deterministic and discrete. We remain in the realm of deontological ethics as long as we do not exceed the computational boundary and stray into analog electronics and physics. On this view, for an agent operating in the physical world and not in a simulation, the most comprehensive agentic boundary is the interface between input sensors and computational representations on the input side, and our output representations leading to actuator mechanisms on the output side. This is as far as we can extend the boundary without taking into account prediction of physical, non-computational consequences. We might call this the *outer agent*.

---
[8] Fisher et al (2013) makes the same point at length, with respect to the deep difficulty of consideration of consequences. Unfortunately, as discussed in the next section, they do not make an equally clear distinction between intentions and actions. Russell et al (2015) retreats subtly from the language of certainty when discussing validation, but misses the opportunity to highlight the distinction made here.

We further observe that an agent might contain computational components that can be independently characterized as agents. We will refer to such a component as an *agentic homunculus*, an agent inside the agent. There might be many different types of agent architectures, in which the homunculi are disjoint, or overlapping, or concentric, for example. There is no requirement that every component of an agent belong to one or more homunculi. As an example, the DG architecture described earlier turns the original outer agent into a homunculus.

Now, suppose that a particular outer agent $A_O$ contains an agentic homunculus $A_H$ the latter taking as input some language $X$ and expressing actions from a language $Y = \{G, B\}$. We assign the element $G$ to mean that the homunculus will choose a Good action and $B$ that it chooses a Bad action. Our deontology is succinct, where $G \equiv H \cap (X^* \times \{G\}^*)$. We stipulate that the outputs of $A_O$ are influenced by the output of $A_H$, without specifying the details.

If the developers of $A_O$ are conscientious, they will of course build $A_H$ so that it always produces the action $G$; in a straightforward implementation it would be easy to prove that $A_H$ complies with the deontology $G$. Subsequent computations in $A_O$ but outside $A_H$ are from the perspective of $A_H$ mere consequences. The developers have created an agent with provably good intentions.

Yet clearly something is amiss. The good intentions are vacuous: the implementation outside $A_H$ is what really determines whether $A_O$ exhibits behavior we would consider Good. On the other extreme, if we specify a deontology for the outer agent $A_O$, we take on a great burden, as discussed in the previous section.

We might imagine some sort of compromise or intermediate solution. Fisher et al (2013) propose exactly that, using a layered architecture. Its "decision apparatus" is an agentic homunculus that incorporates high-level, abstract beliefs, goals, and actions, and leaves the work of perception, interpretation, and action implementation to possibly less structured computational processes. Its deontology is represented via their agent infrastructure layer toolkit (AIL), essentially a semantics specification language. As evidenced by the fact that there is operational software that can verify that deontology, their approach presumably avoids the incomputability issue. It accomplishes this primarily by limiting the representations in the homunculus to abstract logical symbols and including only first-order relationships. Arkin (2008) takes a similar approach, relying on affordances and assuming a discernable ontology of stimuli. Bringsjord et al (2006) work toward formalizing such approaches.

The result is neither vacuous as in $A_H$ above, nor as formidable as producing a specification for the outer agent. However, we might be concerned that this approach looks much like an ethical version of Moravec's paradox (Moravec 1988): such approaches make validation and verification tractable by leaving out the hard part. They analyze the high-level logic of decision-making without consideration of how perceptions are interpreted and actions are executed, not only at the detail level of sensors and actuators but actually eliding substantive, in-the-world semantics entirely. Put another way, it validates and verifies *intentions* rather than actions, treating the latter as consequences outside its purview.

Brundage (2014) makes a similar point:

> This is not merely a quibble with the state of the art that may someday change; rather, it is well-known that even humans make mistakes in conflict situations, and this may be a reflection of the knowledge and computational limitations of finite agents rather than a solvable problem. A combat version of the "frame problem" may apply here: in addition to combatant / non-combatant distinctions, features of situations such as whether a building is a church or a hospital, whether a woman is pregnant, etc. all bear on the consistency of an action with the LOW and ROE [ethical rules] yet are not necessarily amenable to algorithmic resolution by humans or machines…

This is a problem with any layered architecture. The relationship between logical intentions and actions is mediated by semantics, models of the world, plans, and commitment. A logical term cannot refer to the real world purely logically - there must be an ostensive grounding. A model of the world is incomplete and usually includes theoretical terms dependent on a complex fabric of relationships among terms, both logical and ostensive. A plan can be purely logical, but is also completely dependent on semantics and models. Finally, an agent can conceivably change its intentions during the course of executing on them, rendering its commitment incomplete.

Take for example a system that has the mandate "never kill a human" expressed in some logical form. Let us stipulate that the system is able to distinguish humans from non-humans successfully. What does it mean to "kill"?  The semantics must be elaborated in some sort of model within the agent. Suppose that such a model elaborates various methods of killing and their associated instruments, such as weapons or poisons. We can be confident that such a taxonomy can never be complete. If the agent gains reward through killing a human, it will find a method that does not violate its fixed set of rules. If, in contrast, we attempt to characterize the action abstractly, such as "do not be the cause of a human's death," then we have deferred the problem to the word "cause," which is fraught with other difficulties, such as assignment of responsibility or risk.

*These concerns are not superficial, they are fundamental.* The semantics of the world cannot be captured with purely abstract symbols unless the environment is highly constrained. Logical reasoning about intentions within a homunculus is not a substitute for rules of action for an outer agent. There is a fairly direct tradeoff between the extent to which intentions connect to actions and the complexity of the model of intention.

To summarize, a limited and abstract approach to verification and validation can be useful for agentic and autonomous systems operating in a narrow and regulated domain; but for general artificial intelligence, which by definition needs to be able to operate in any environment, logical intentions are the tip of the ethical iceberg. Numerous catastrophic scenarios have been elaborated in which artificial intelligence destroys humanity inadvertently. Using language of certainty with respect to intentions belies the vast underdetermination of resulting actions and then consequences.

**<u>Discussion</u>**

We have explored a very general formalism describing computational agents and a way to classify their behavior as Good or Bad. In doing so we have identified extensive limitations and challenges in both verification and validation of such agents:

- A general automated procedure cannot exist to verify an agent's possible actions against an explicit account of which actions are considered Good.
- Manual verification proofs are extremely arduous to produce, and to ensure safety, must be created for every revision of an agent's software.
- Specifying a set of rules of behavior also faces computability issues, particularly in regard to an agent that has any ability to self-improve or create successors.
- Validating a specification is difficult for any kind of software; for an agent operating in the physical world, it would require a complete and analytic physical model of the world so as to predict consequences.
- Layered agent architectures that rely on a logic-based "homunculus" with tractable validation and verification offer an illusion of progress, but fail to solve the real problem because they only address ungrounded intentions, not actions or consequences.
- Most proposed solutions to these types of problems either impose strong limits on the capabilities of the agent, gloss over uncertainties associated with connecting logical reasoning to real-world perceptions and actions, or both.

There is apparently no way out of this tower of difficulties. It might be possible to thread the needle of verification by finding a way to specify a computable deontology that nevertheless does not reduce the capabilities of the agent below those of general intelligence. Then our proposed "computational superego," or in the alternative, a lengthy, labor-intensive iterative theorem proving effort, could be implemented. But such a hard-earned success in verification provides only limited comfort, much less certainty, that the outcomes of the agent's actions will be desirable.

We reiterate that this analysis is aimed at *general* artificial intelligence and questions of *certainty.* Some approaches mentioned here may prove useful in *narrow* artificial intelligence applications and domains. Similarly, they may be able to provide some *confidence* that one approach to general artificial intelligence is safer than another. Nevertheless, the results and illustrations here show that efforts to achieve even those more modest goals are likely limited.

This paper has demonstrated that in building general artificial intelligence, we cannot achieve certainty or proof that its outcomes will be Good, Safe, Beneficial, or Friendly. Consequently, we should stop talking about it as though that is a possibility. Though we can improve probabilities and carefully consider safety in our designs, to a significant extent we will simply have to take our chances. Analyzing risk scenarios by comparison to a standard of perfect safety misdirects our focus and efforts away from that which we can actually control.

**Acknowledgements and Affiliations**

This work was supported in part by the Future of Life Institute (futureoflife.org) FLI-RFP-AI1 program, grant #2015-144585, through an affiliation with Theiss Research, La Jolla, CA 92037 USA. I am a minority shareholder in and occasional consultant to eCortex, Inc., a small research company.  I thank Samuel Arbesman and Kristin Lindquist for helpful comments on readability, and Seth Herd for numerous discussions on these topics.